\documentclass[]{academic_template}

\usepackage[toc,page,header]{appendix}

\usepackage{amssymb}
\usepackage{amsmath}
\usepackage{wrapfig}

\usepackage{makecell}
\usepackage{algorithm}      
\usepackage{algpseudocode}  

\usepackage{graphicx}

\usepackage{listings}
\usepackage{url}
\usepackage{booktabs}
\usepackage{wrapfig}

\usepackage{colortbl}
\usepackage{xcolor}

\usepackage{caption}
\usepackage{subcaption}

\setlogoheight{14mm}   
\setlogospacing{5mm}
\setsjtublue 

\settitlerulethickness{3pt}  

\abstractboxon 
\setabstractframecolor{gray} 
\setabstractbgcolor{gray!10} 
\setlogotolineshift{5mm} 

\settoprulethickness{2.5pt}  
\setbottomrulethickness{1.5pt}

\title{Partial Weakly-Supervised Oriented Object Detection}

\author[1]{Mingxin Liu}
\author[2]{Peiyuan Zhang}
\author[1]{Yuan Liu}
\author[1]{Wei Zhang}
\author[3]{Yue Zhou}
\author[1]{Ning Liao}
\author[1]{Ziyang Gong}
\author[2]{Junwei Luo}
\author[4]{Zhirui Wang}
\author[5, \dagger]{Yi Yu}
\author[1, \dagger]{Xue Yang}

\affiliation[1]{Shanghai Jiao Tong University}
\affiliation[2]{Wuhan University}
\affiliation[3]{East China Normal University}
\affiliation[4]{Aerospace Information Research Institute}
\affiliation[5]{Southeast University}

\contribution[\dagger]{Corresponding Author}

\abstract{
The growing demand for oriented object detection (OOD) across various domains has driven significant research in this area. However, the high cost of dataset annotation remains a major concern. Current mainstream OOD algorithms can be mainly categorized into three types: (1) fully supervised methods using complete oriented bounding box (OBB) annotations, (2) semi-supervised methods using partial OBB annotations, and (3) weakly supervised methods using weak annotations such as horizontal boxes or points. However, these algorithms inevitably increase the cost of models in terms of annotation speed or annotation cost. To address this issue, we propose: (1) \textbf{the first Partial Weakly-Supervised Oriented Object Detection (PWOOD) framework} based on partially weak annotations (horizontal boxes or single points), which can efficiently leverage large amounts of unlabeled data, significantly outperforming weakly supervised algorithms trained with partially weak annotations, also offers a lower cost solution; (2) \textbf{Orientation-and-Scale-aware Student (OS-Student)} model capable of learning orientation and scale information with only a small amount of orientation-agnostic or scale-agnostic weak annotations; and (3) \textbf{Class-Agnostic Pseudo-Label Filtering strategy (CPF)} to reduce the model's sensitivity to static filtering thresholds. Comprehensive experiments on DOTA-v1.0/v1.5/v2.0 and DIOR datasets demonstrate that our PWOOD framework performs comparably to, or even surpasses traditional semi-supervised algorithms. Our code will be made publicly available.
}

\date{\today}

\checkdata[Code]{\url{https://github.com/VisionXLab/PWOOD}}
\begin{document}
\maketitle


\section{Introduction}
\label{sec:intro}

\begin{wrapfigure}{r}{0.5\textwidth}
    \centering
    \includegraphics[width=\linewidth]{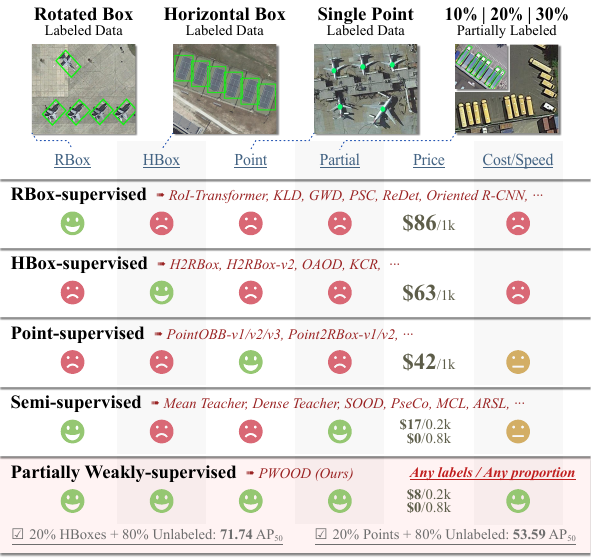}
    \caption{The main paradigmatic types of existing oriented object detection. 
    Compared to these settings, our proposed Partial Weakly-supervised PWOOD offers high efficiency, low costs, and good performance (accuracy on DOTA-v1.0 is shown).}
    \label{fig:setting}
\end{wrapfigure}

In the field of oriented object detection, fully supervised learning~\cite{ding2019learning,yang2019scrdet,yang2021r3det,xie2021oriented} based on rotated bounding box annotations has been dominant, as shown in Figure~\ref{fig:setting}(a). However, obtaining rotated bounding box annotations for a large number of images is extremely costly and labor-intensive, which poses a significant challenge for model training. 

To address the issue of the difficulty in obtaining a large amount of data, semi-supervised oriented object detection (SOOD)~\cite{hua2023sood,wang2024multi} algorithms have been proposed. As shown in Figure\ref{fig:setting}(d), these algorithms suggest using only a small amount of labeled data within the dataset and making full use of unlabeled data through the application of a self-training framework~\cite{li2022pseco,liu2021unbiased,liu2022unbiased,wang2022consistent}, thereby enhancing the performance of the detector under the condition of small-batch labeled data. Moreover, weakly supervised oriented object detection (WOOD)~\cite{yu2024point2rbox,luo2024pointobb, yang2023h2rbox,yu2023h2rboxv2} offers another cost-effective solution. Notably, methods such as H2RBox-v2~\cite{yu2023h2rboxv2} and Point2RBox-v2~\cite{yu2025point2rbox} have demonstrated the feasibility of training detectors using horizontal bounding box annotations and single point annotations, as illustrated in Figure~\ref{fig:setting}(b-c).

To further reduce annotation costs and fully leverage weakly annotated and unlabeled data, we propose a new oriented object detection framework, named Partial Weakly-Supervised Oriented Object Detection (PWOOD), that utilizes only a subset of weakly annotated data (e.g. horizontal bounding box or single point), as demonstrated in Figure~\ref{fig:setting}(e). Inheriting the teacher-student paradigm~\cite{tarvainen2017mean}, we introduce orientation learning and scale learning strategies to address the limitation of the teacher-student framework in learning orientation and scale information from partially weak annotations. This enables the student model to acquire the ability to learn the precise pose of the object, thereby achieving an Orientation-and-Scale-aware Student (OS-Student) model. However, a significant limitation of teacher-student paradigm is their reliance on static thresholds for pseudo-label selection~\cite{wang2024multi, liu2021unbiased}. This approach often leads to threshold inconsistency, which can adversely affect the robustness and generalization ability of the model~\cite{wang2022consistent, wang2022freematch,zhong2020boosting}. 
To tackle this challenge, we design a Class-Agnostic Pseudo-Label Filtering (CPF) based on a Gaussian Mixture Model. By employing maximum likelihood estimation, we dynamically adjust the filtering threshold, allowing the model to adaptively generate pseudo-labels that are more stable and aligned with the teacher's performance.  We apply this framework to a setting with partially annotated horizontal bounding boxes, achieving comparable performance to the traditional semi-supervised baseline. Additionally, we also validate the framework on datasets with partial point annotations.

Inspired by the teacher-student paradigm, we leverage a small amount of weakly-annotated data for pre-training, enabling the teacher model to generate pseudo-labels for unlabeled data. These pseudo-labels are then utilized to train the student model, allowing the student to learn from both the limited weakly-annotated data and abundant unlabeled data. As a result, the quality of pseudo-labels and the strategy employed for filtering them are pivotal to the overall performance of the model. However, a significant limitation of existing methods is their reliance on static thresholds for pseudo-label selection~\cite{wang2024multi, liu2021unbiased}. This approach often leads to threshold inconsistency, which can adversely affect the robustness and generalization ability of the model~\cite{wang2022consistent, wang2022freematch,zhong2020boosting, chen2022dense}. To tackle this challenge, we focus on enhancing the quality of pseudo-labels produced by the teacher model. We design a Class-Agnostic Pseudo-Label Filtering (CPF) based on a Gaussian Mixture Model. By employing maximum likelihood estimation, we dynamically adjust the filtering threshold, allowing the model to adaptively generate pseudo-labels that are more stable and aligned with the teacher's performance. This approach not only mitigates the issue of threshold inconsistency but also improves the model's ability to handle diverse and complex scenarios, ultimately leading to more robust detection performance.

The contributions of this work are as follows:
\begin{itemize}

\item  To our best knowledge, we propose the first Partial Weakly-Supervised Oriented Object Detection (PWOOD) framework, aiming to achieve competitive performance in a cost-conscious setting.

\item We utilize partially weak annotations to enable the student model to learn orientation and scale information, resulting in an orientation-and-scale-aware student.

\item Class-Agnostic Pseudo-Label Filtering is introduced for teacher models to address threshold inconsistency and reduce sensitivity to static thresholds, thereby enhancing the robustness of the model.

\item Extensive experiments on DOTA-v1.0/v1.5/v2.0 and DIOR datasets show that our framework exhibits performance comparable to SOOD algorithms that rely on partially annotated rotated bounding boxes, thereby achieving superior results with weaker supervision.

\item  We generalize the PWOOD framework to support various annotation forms. Preliminary experiments show it reduces gaps between different annotations, offering cost-effective training options for detectors.

\end{itemize}

\section{Related Work}
\label{sec:formatting}

\textbf{Semi-supervised Oriented Object Detection:} In semi-supervised object detection algorithms, the teacher-student framework is a common paradigm~\cite{li2022rethinking, liu2023mixteacher, nie2023adapting, sun2021makes}, where the teacher model generates pseudo-labels from unlabeled data to supervise the training of the student model, and the student model updates the teacher model using exponential moving average (EMA)
. For instance, SOOD~\cite{hua2023sood} leverages optimal transport theory to define a cost matrix, calculating the distance between pseudo-labels and predictions. Meanwhile, MCL~\cite{wang2024multi} builds upon the teacher-student paradigm by introducing Gaussian centerness into the label assignment strategy and incorporating adaptive label assignment for unsupervised learning based on the characteristics of different feature layers. However, both models rely on rotated bounding box annotations for training, which are difficult to obtain in practice and come with high annotation costs, significantly increasing the overall training cost.

\textbf{HBox-supervised oriented object detection:} Weakly supervised object detection algorithms~\cite{zhang2021weakly, yang2019towards,bilen2015weakly, iqbal2021leveraging, zhu2023knowledge} aim to train detectors using more accessible and cost-effective annotations. 
Among them, numerous methods are based on horizontal bounding box annotations~\cite{ wang2024explicit,zhu2023knowledge,tian2021boxinst, sun2021oriented,li2022box}. For instance, H2RBox~\cite{yang2023h2rbox} leverages horizontally annotated data and employs angle consistency loss, enabling the detector to learn orientation information. Based on H2RBox, H2RBox-v2~\cite{yu2023h2rboxv2} introduces symmetric learning. In addition to random rotation augmentation, it incorporates vertical flipping and self-supervised symmetry learning, enriching the learning pathways for orientation information. Moreover, under horizontal bounding box supervision, AFWS~\cite{lu2024afws} proposes an angle-free approach that decouples horizontal information from rotational information using concentric circles, simplifying the training process of the model.

\textbf{Point-supervised oriented object detection:} In recent studies, multiple approaches have been developed for oriented object detection using point supervision~\cite{chen2021points, chen2022point, he2023learning,ying2023mapping}. For instance, P2RBox~\cite{cao2023p2rbox}, PMHO~\cite{zhang2024pmho}utilize SAM's~\cite{kirillov2023segment}  zero-shot point-to-mask functionality to perform detection with point-based prompts. Another method, Point2RBox~\cite{yu2024point2rbox}, adopts an end-to-end strategy by integrating diverse knowledge sources. Additionally, PointOBB~\cite{luo2024pointobb} introduces a technique for generating rotated bounding boxes from points, employing scale-sensitive consistency and multiple instance learning. Building on this, PointOBB-v2~\cite{yu2025point2rbox} enhances the process by constructing a class probability map and applying principal component analysis to produce pseudo RBox annotations, further pushing the boundaries of point-supervised detection. 
Point2RBox-v2~\cite{yu2025point2rbox}  incorporates novel losses based on Gaussian overlap and Voronoi tessellation to enforce spatial layout constraints, along with additional modules such as edge loss, consistency loss, and copy-paste augmentation to further enhance its effectiveness. 
\section{Method}

\begin{figure*}[t]
    \centering
    \includegraphics[width=0.99\linewidth]{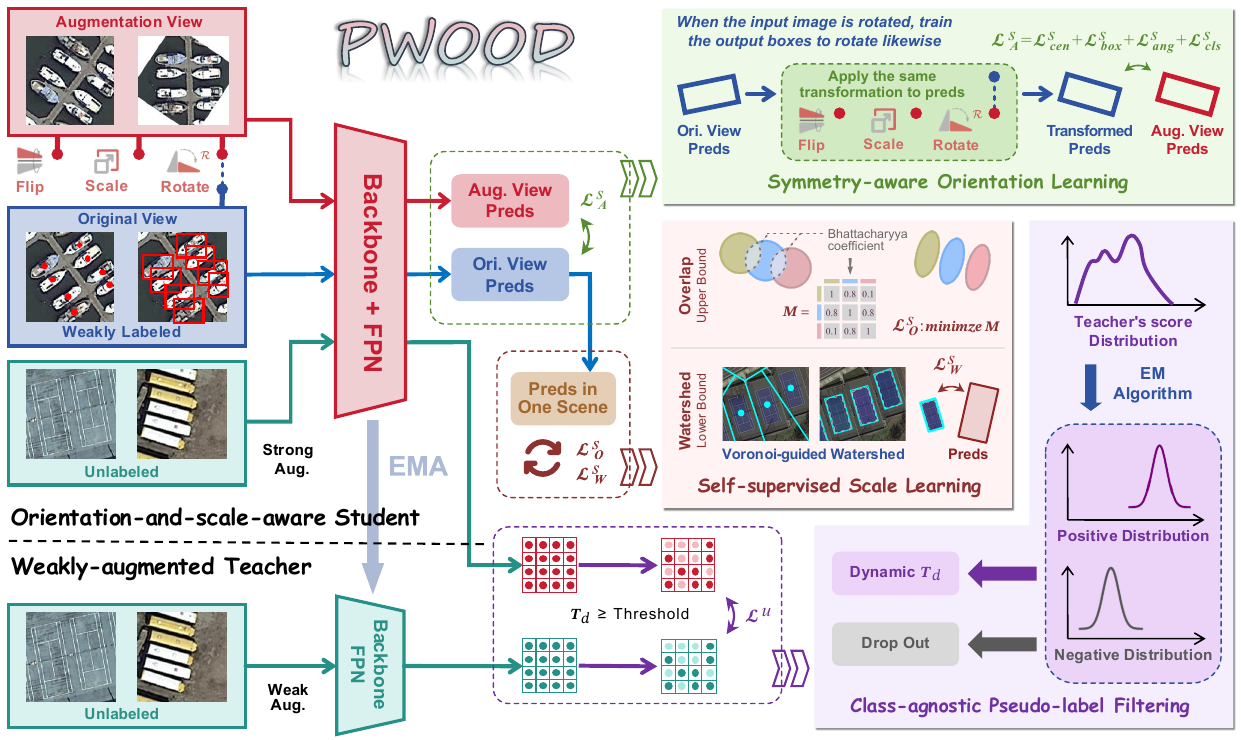}
    \caption{The overview of the proposed PWOOD framework. Orientation Learning and Scale Learning modules enables the Orientation-and-Scale-aware Student to learn both scale and orientation information from weakly annotated data, as well as Class-Agnostic Pseudo-Label Filtering mechanism enhances pseudo-label quality by leveraging dynamic thresholds.}
    \label{fig:main}
\end{figure*}

In this section, we elaborate on how to leverage orientation learning and scale learning strategies to form an Orientation-and-Scale-aware Student (OS-Student) and construct the PWOOD framework, achieving oriented object detection under the supervision of a small amount of weakly annotated data. First, we introduce our proposed PWOOD framework and its working mechanism. Next, we present the OS-Student model trained with scale learning and orientation learning modules. Additionally, to address the inconsistency in pseudo-label assignment thresholds for semi-supervised learning, we propose a Class-Agnostic Pseudo-Label Filtering (CPF) . It ensures more reliable pseudo-label generation and enhances the overall robustness of the model. Finally, we delineate the composition of the overall loss function for the proposed model. 

\subsection{PWOOD Framework}\label{sec:3_1}
Given the limited scale of weakly-annotated data, to fully exploit the potential of unlabeled data and further enhance the performance of the OS-Student model, we utilize the teacher model to generate pseudo-labels~\cite{sohn2020simple} for unlabeled data. As illustrated in Figure~\ref{fig:main}, both the teacher and the OS-Student share identical architectures in their backbone, neck, and head components~\cite{tarvainen2017mean,chen2022dense}. During the pre-training phase, the weakly-annotated data is utilized to train the OS-Student. The OS-Student automatically learns the scale and orientation information from a small amount of weakly annotated data through the Orientation Learning and Scale Learning modules depicted in Figure~\ref{fig:main}. When the model reaches the burn-in step, the weights of the OS-Student are mirrored to the teacher. Subsequently, unlabeled data is introduced, and the model training enters the burn-in stage.

During the burn-in stage, the data undergoes weak and strong augmentation before being fed into the teacher and OS-Student networks~\cite{hua2023sood}, respectively, to obtain predictions. Based on the teacher's confidence scores for each pseudo-box and a Gaussian Mixture Model, we employ a Class-Agnostic Pseudo-Label Filtering to filter the pseudo-boxes, thereby generating high-quality pseudo-labels. This approach enhances the robustness and accuracy of the model in leveraging both weakly-annotated and unlabeled data for improved object detection performance.

During this process, the student model is not only trained using pseudo-labels but also updates the weights of the teacher model through an Exponential Moving Average (EMA)~\cite{tarvainen2017mean} approach. This dynamic weight update mechanism allows the orientation-and-scale-aware capabilities learned by the student to be effectively transferred back to the teacher model, creating a positive feedback loop. As training progresses, the accuracy of the pseudo-labels generated by the teacher model gradually improves, further enhancing the ability of the OS-Student.

\subsection{OS-Student}\label{sec:3_2}

\textbf{Orientation Learning}:
To enable the student model to learn orientation information, we introduce symmetry learning~\cite{yu2023h2rboxv2}. During training, the input images are vertically flipped or randomly rotated to generate transformed views. These views are fed into the network to obtain predictions for both the original and transformed images. The labeled data undergoes the same transformations, forming weakly-supervised pairs. Additionally, since there is a deterministic mapping relationship between the original image and its transformed views, the predictions of the model for the original and transformed views should satisfy the same mapping relationship, provided the prediction accuracy is high. Based on this principle, the predictions of the network for the original and transformed views form self-supervised pairs. Through the self-supervised and weakly supervised branches, the student model gains the ability to learn orientation information in a weakly annotated setting. We formulate an angle loss $\mathcal{L}^s_\text{Ang}$ to ensure that the OS-Student can effectively learn orientation information from weakly annotated horizontal bounding boxes:
\begin{equation}
    \mathcal{L}^s_\text{Ang} = 
\begin{cases}
L^s_\text{Ang}(\theta_{flp} + \theta, 0) & trans  = \text{flip} \\
L^s_\text{Ang}(\theta_{rot} - \theta, \mathcal{R}) & {trans = \text{rotate($\theta$)} }.
\end{cases} 
\end{equation}

The computation of the angle loss is related to the image transformation method $trans$, which involves either vertical flipping or random rotation by an angle $\theta$. $L^s_\text{Ang}$ denotes Smooth-L1 loss.

\textbf{Scale Learning:} Considering the presence of even weaker annotation forms in the dataset, such as single point annotations, which lack scale information, we introduce a scale learning strategy to enable the student model to learn scale information effectively under such weakly annotated conditions. We employ spatial layout learning~\cite{yu2025point2rbox} to guide the model's scale prediction accuracy by estimating upper and lower bounds on object scales.

To obtain the upper bound of the object scale, we treat the oriented bounding box as a Gaussian distribution and introduce the Bhattacharyya coefficient~\cite{yang2023detecting} to measure the overlap between Gaussian distributions. We find the upper bound by minimizing the Gaussian overlap between different pre-
dicted bounding boxes. Consequently, we derive the Gaussian overlap loss as follows:
\begin{equation}
    \mathcal{L}_O^s = \frac{1}{N}\sum_{\substack{i, j = 1, \ i\neq j}}^N B\left ( \mathcal{N}_i, \mathcal{N}_j \right ),
\end{equation}
where $N$ denotes the number of predicted bounding boxes, $\mathcal{N}_i$ is the Gaussian distribution of the $i$-th box, and $B\left ( \mathcal{N}_i, \mathcal{N}_j \right )$ denotes the Bhattacharyya coefficient between the $i$-th and $j$-th predicted bounding boxes.

Moreover, to obtain the lower bound of the object scale, the Voronoi diagram~\cite{aurenhammer1991voronoi} and the watershed algorithm~\cite{vincent1991watersheds} are introduced to calculate Voronoi Watershed Loss. Using the ridges of the Voronoi diagram as background markers and point annotations as foreground markers, we apply the watershed algorithm to segment the image and obtain the basin regions for each object. By rotating these watershed regions to align with the current predicted orientation, we calculate the regression objects for width and height. Finally, we use the Gaussian Wasserstein Distance loss~\cite{yang2023detecting} to regress the width $w_t$ and height $h_t$ of the objects:
\begin{equation}
    \mathcal{L}_W^s = L_\text{GWD}\!\left (\! \begin{bmatrix}
 w/2 & 0 \\
 0 & h/2
\end{bmatrix} ^{2}\!,\begin{bmatrix}
 w_t/2 & 0 \\
 0 & h_t/2
\end{bmatrix} ^{2} \right ).
\end{equation}

Then, we introduce class loss $\mathcal{L}^s_{cls}$, centerness loss $\mathcal{L}^s_{cen}$, box loss $\mathcal{L}^s_{box}$, respectively. And the supervised loss $\mathcal{L}^s$ of OS-Student is as follows:
\begin{equation}
    \begin{aligned}
    \mathcal{L}^s = &\alpha_1  \mathcal L^s_{cls}(p_{(x,y)}, c_{(x,y)}) + \alpha_2 \mathcal L^s_{cen}(cn'_{(x,y)}, cn_{(x,y)}) \\ &\quad + 
     \alpha_3  \mathcal L_{box}^s(pr_{(x,y)}, gt_{(x,y)}) + 
    \alpha_4\mathcal{L}^s_\text{Ang} \\
    &\quad+ \alpha_5\mathcal{L}_O^s + \alpha_6\mathcal{L}_W^s.
    \end{aligned}
\end{equation}

According to FCOS~\cite{tian2019fcos},  each positive point $(x,y)$ on the feature map corresponds to a potential object center or anchor point in the input image. $p$, $cn'$, $pr$ represent the category score, centerness, and rotated bounding box predictions of the OS-Student, while $c$, $cn$, $gt$ are the corresponding weak ground truth labels. $\mathcal L^s_{cls}$, $\mathcal L^s_{cen}$ and $\mathcal L_{box}^s$ denotes focal loss~\cite{lin2017focal}, cross entropy loss, and IoU loss~\cite{yu2016unitbox}, respectively. 
For hyperparameters, $\alpha_1$, $\alpha_2$, $\alpha_3$ are all set to 1, following FCOS\cite{tian2019fcos} where this configuration was shown effective. $(\alpha_4, \alpha_5, \alpha_6)$ is set to $(0.2, 10, 5)$ by default, ablations are provided in the supplementary materials.

\subsection{Class-Agnostic Pseudo-Label Filtering}\label{sec:3_3}

The quality of pseudo-labels generated by the teacher model is a critical factor influencing the overall performance of the detection framework. Most existing approaches~\cite{wang2024multi, xu2021end, liu2021unbiased, sohn2020simple} adopt a static thresholding strategy to filter pseudo-labels, which is typically determined empirically and lacks adaptability to the inherent characteristics of the data and the dynamic distribution of pseudo-labels across different training stages. Specifically, during the initial phases of training, the teacher model tends to produce pseudo-labels with relatively low confidence scores due to its underdeveloped representation capability, while as training progresses, the teacher model gradually improves, leading to higher-quality pseudo-labels with more reliable confidence estimates. And we find that during the training stage, the model exhibits high sensitivity to threshold settings.

The issue above highlights the need for a more adaptive and data-driven approach~\cite{wang2022freematch,zhong2020boosting} to pseudo-label selection in semi-supervised object detection. To address this, we propose Class-Agnostic Pseudo-Label Filtering (CPF). Based on a Gaussian Mixture Model~\cite{wang2022consistent, zhao2019image}, we model the scores of pseudo boxes generated by the teacher as a mixed model  \(\mathcal{P}(s)\) of two one-dimensional Gaussian distributions:
\begin{equation}
    \mathcal{P}(s) = w_p\mathcal{N}_p(\mu_p, (\sigma_p)^2) + w_n\mathcal{N}_n(\mu_n, (\sigma_n)^2) ,
\end{equation}
where $\mathcal{N}_p$, $\mathcal{N}_n$ denote the distributions of positive samples and negative samples, respectively. The initial mean of the positive sample distribution $\mu_p^{(0)}$ is set to the maximum of the predicted score $s$, while the negative value $\mu_n^{(0)}$ is set to the lowest predicted score. The weights of the two one-dimensional Gaussian distributions $w_p^{(0)}$ and $w_n^{(0)}$ in the mixture model are both initialized to 0.5. Then, we employ the expectation-maximization (EM) algorithm to deduce the posterior $\mathcal{P}_p$, which  represents the likelihood that a detection ought to be designated as the pseudo-object for the student:
\begin{equation}
    T_d = \underset{s}{\text{argmax}} \mathcal{P}_p(s, \mu_p, (\sigma_p)^2).
\end{equation}

\subsection{Overall Loss}
\label{sec:3_4}

In our proposed PWOOD framework, the overall loss function of the model consists of two distinct components: the supervised loss $\mathcal{L}^s$ within the OS-Student and the unsupervised loss $\mathcal{L}^u$ derived from the pseudo-labels generated by the teacher model to guide the learning of the OS-Student. We elaborate on the former in Section \ref{sec:3_2}, and the latter is defined as follows:
\begin{equation}
    \begin{aligned}
    \mathcal{L}^u = \omega & (\mathcal{L}^u_{cls}(\mathcal{T}^c_{(x,y)}, \mathcal{S}^c_{(x,y)}) + \mathcal{L}^u_{cen}(\mathcal{T}^{cen}_{(x,y)}, \mathcal{S}^{cen}_{(x,y)}) \\
    &\quad + \mathcal{L}^u_{box}(\mathcal{T}^{logit}_{(x,y)}, \mathcal{S}^{logit}_{(x,y)})),
    \end{aligned}  
\end{equation}
the predictions of the teacher and student models, denoted as $\mathcal{T}$ and $\mathcal{S}$, are represented by the tuple $(c, cn, logit)$ where $c$ is the class score, indicating the confidence of the predicted class, $cn$ is the predicted center-ness, measuring how close the point is to the center of the object, $logit$ represents the distances from the point to the predicted bounding box's left, top, right, and bottom boundaries. $\mathcal{L}^u_{cls}$ and $\mathcal{L}^u_{cen}$ are binary cross-entropy losses, used for classification and centerness prediction, respectively. $\mathcal{L}^u_{box}$ is the Smooth L1 loss, applied to the regression of the bounding box distances. The weight $\omega$ of each loss component is associated with the score of each point, ensuring that points with higher confidence contribute more to the overall loss. This weighting mechanism helps the model focus on high-quality predictions, improving both localization and classification accuracy. The overall loss of PWOOD is as 
follows:
\begin{equation}
    \mathcal{L} = \alpha\mathcal{L}^s +  \beta\mathcal{L}^u,
\end{equation}
Both $\alpha$ and $\beta$ are set to 1, ablations are in the supplement.

\section{Experiment}
\begin{table*}[t]
\centering
 \vspace{-5pt}
    \caption{mAP comparison on DIOR \textit{val set}, DOTA-v1.0 \textit{val set} and DOTA-v2.0 \textit{val set}.}  
    \label{tab_dior}
 \vspace{-5pt}
  \resizebox{0.98\textwidth}{!}{
  \begin{tabular}{c l c c c c c c c c c}
  \toprule
        \multirow{2}{*}{\textbf{Task}} & \multirow{2}{*}{\textbf{Method}} & \multicolumn{3}{c}{\textbf{DIOR}} & \multicolumn{3}{c}{\textbf{DOTA-v1.0}} & \multicolumn{3}{c}{\textbf{DOTA-v2.0}} \\
        \cmidrule(r){3 - 5}  \cmidrule(r){6 - 8} \cmidrule(r){9 - 11} 
         & & 10\% & 20\% & 30\%  & 10\% & 20\% & 30\% & 10\% & 20\% & 30\% \\
        \midrule
        \multirow{2}{*}{WOOD}
          & H2RBox-v2~\cite{yu2023h2rboxv2} & 44.54 & 51.33 & 53.45 & 47.96 & 54.38 & 58.65 & 19.07 & 28.56 & 31.73 \\
            & Point2RBox-v2~\cite{yu2025point2rbox}  & 28.77  & 32.81  & 33.07  & 35.24 & 40.39 & 45.09 & 12.18  & 16.74 & 16.62 \\
        \midrule
        SOOD & Vanilla Baseline (\textit{w}/ Partial RBox) & 54.01	& 57.07	& 60.25 & 56.03 & 62.82 & 64.97 & 24.77 & 34.03 & 37.30 \\
        \hline
        \multirow{2}{*}{\parbox{1.5cm}{\small \centering \textbf{PWOOD\\(ours)}}} 
        & \cellcolor{gray!20} \textit{w}/ Partial HBox Annotations & \cellcolor{gray!20} \textbf{54.33} & \cellcolor{gray!20} \textbf{57.89} & \cellcolor{gray!20} \textbf{60.42} & \cellcolor{gray!20} \textbf{56.92} & \cellcolor{gray!20} \textbf{62.93} & \cellcolor{gray!20} \textbf{65.42}  & \cellcolor{gray!20} \textbf{31.03} & \cellcolor{gray!20}  \textbf{36.39} & \cellcolor{gray!20} \textbf{40.27} \\
        & \cellcolor{gray!20} \textit{w}/ Partial Point Annotations  & \cellcolor{gray!20} 32.04 & \cellcolor{gray!20} 35.17 & \cellcolor{gray!20} 36.44 & \cellcolor{gray!20} 42.35 & \cellcolor{gray!20} 45.01 & \cellcolor{gray!20} 49.12 & \cellcolor{gray!20} 13.44 & \cellcolor{gray!20} 18.49 & \cellcolor{gray!20} 23.85 \\
        \bottomrule
    \end{tabular}}
   \vspace{-5pt}
\end{table*}

Our experiments are carried out on the oriention detection tool kits: MMRotate 0.3.4~\cite{zhou2022mmrotate} and Pytorch 1.13.1~\cite{paszke2019pytorch}.

\subsection{Datasets and Experimental Setting}
\textbf{DOTA Dataset-Partial:} We conducted experiments on all three versions of the DOTA dataset~\cite{xia2018dota}. DOTA-v1.0 contains 188,282 instances across 15 categories, while DOTA-v1.5/2.0 adds numerous small object annotations
Following the SOOD~\cite{hua2023sood} method, we randomly sampled 10\%, 20\%, and 30\% of the images from the DOTA training set to serve as annotated data. 
During the data loading process, these annotated images underwent specific transformations to remove orientation and scale information, resulting in semi-weakly annotated data. The remaining images are treated as unannotated data. For validation purposes, the DOTA-v1.5 validation set is employed.

\textbf{DIOR Dataset-Partial:} DIOR~\cite{li2020object} comprises a total of 23,463 images, with 11,725 images allocated for training and 11,738 images for testing. 
To facilitate our experiments, we have offline partitioned the DIOR training set into subsets containing 10\%, 20\%, and 30\% of the data as supervised datasets, while the remaining data is treated as unsupervised. The DIOR test set is utilized for both validation and testing purposes, ensuring a comprehensive evaluation of the model's performance.

\textbf{Experimental Setting:} We adopt the FCOS~\cite{tian2019fcos} detector with ResNet50~\cite{he2016deep} backbone and FPN~\cite{lin2017feature} neck. 
We employ a simplified version of the MCL\footnote{Some of MCL's components lack universality. For instance, GCA cannot be deployed in Point2Rbox-v2, which is independent of centerness.}, excluding the GCA and CCSL modules~\cite{wang2024multi}, as our SOOD baseline, referred to as the Vanilla Baseline. 
In addition, we further compare the performance of our proposed PWOOD framework with weakly supervised detectors~\cite{yu2025point2rbox,yu2023h2rboxv2} trained on partially weakly annotated data to ensure the comprehensiveness and integrity of experiments. To ensure a fair and comprehensive evaluation, we adopt Average Precision (AP) as the primary metric for benchmarking against existing literature. In the experiments, the PWOOD model is trained for 180$k$ iterations under the 30\% partial and full settings, while for the 10\% and 20\% partial settings, it is trained for 120$k$ iterations~\cite{wang2024multi}. All models are trained using the AdamW optimizer~\cite{loshchilov2017decoupled}. 

\begin{table*}[t]
\fontsize{8.5pt}{10pt}\selectfont
\setlength{\tabcolsep}{1.9mm}
\setlength{\aboverulesep}{0.4ex}
\setlength{\belowrulesep}{0.4ex}
\setlength{\abovecaptionskip}{1.5mm}
\centering
\caption{Detection performance of each category on the DOTA-v1.0 \textit{test set} and the mean AP$_\text{50}$ of all categories.}
\label{tab:exp_dotav1}
\resizebox{\textwidth}{!}{
\begin{tabular}{l|ccccccccccccccc|c}
\toprule
\textbf{Methods}  &  \textbf{PL}$^1$    & \textbf{BD}    & \textbf{BR}    & \textbf{GTF}   & \textbf{SV}    & \textbf{LV}    & \textbf{SH}    & \textbf{TC}    & \textbf{BC}    & \textbf{ST}    & \textbf{SBF}   & \textbf{RA}    & \textbf{HA}    & \textbf{SP}    & \textbf{HC}    & \textbf{AP}$_\text{50}$  \\ \hline
\rowcolor{gray!20} \multicolumn{17}{l}{$\blacktriangledown$ \textit{RBox-supervised OOD}} \\ \hline
RepPoints (2019) ~\cite{yang2019reppoints} & 86.7  & 81.1  & 41.6  & 62.0  & 76.2  & 56.3  & 75.7  & 90.7  & 80.8  & 85.3  & 63.3 & 66.6  & 59.1  & 67.6  & 33.7  & 68.45 \\ 
RetinaNet (2017) ~\cite{lin2017focal}  & 88.2  & 77.0  & 45.0  & 69.4  & 71.5  & 59.0  & 74.5  & 90.8  & 84.9  & 79.3  & 57.3 & 64.7  & 62.7  & 66.5  & 39.6  & 68.69 \\
GWD (2021) ~\cite{yang2021rethinking}  & 89.3  & 75.4  & 47.8  & 61.9  & 79.5  & 73.8  & 86.1  & 90.9  & 84.5  & 79.4  & 55.9 & 59.7  & 63.2  & 71.0  & 45.4  & 71.66 \\
FCOS (2019) ~\cite{tian2019fcos}  & 89.1  & 76.9  & 50.1  & 63.2  & 79.8  & 79.8  & 87.1  & 90.4  & 80.8  & 84.6  & 59.7 & 66.3  & 65.8  & 71.3  & 41.7  & 72.44 \\
S$^2$A-Net (2022) ~\cite{han2022align}  & 89.2  & 83.0  & 52.5  & 74.6  & 78.8  & 79.2  & 87.5  & 90.9  & 84.9  & 84.8  & 61.9 & 68.0  & 70.7  & 71.4  & 59.8  & 75.81 \\ \hline
\rowcolor{gray!20} \multicolumn{17}{l}{$\blacktriangledown$ \textit{HBox-supervised OOD}} \\ \hline
H2RBox (2023) ~\cite{yang2023h2rbox}  & 88.5  & 73.5  & 40.8  & 56.9  & 77.5  & 65.4  & 77.9  & 90.9  & 83.2  & 85.3  & 55.3 & 62.9  & 52.4  & 63.6  & 43.3  & 67.82 \\
EIE-Det (2024) ~\cite{wang2024explicit}   & 87.7 & 70.2 & 41.5 & 60.5 & 80.7 & 76.3 & 86.3 & 90.9 & 82.6 & 84.7 & 53.1 & 64.5 & 58.1 & 70.4 & 43.8 & 70.10 \\
H2RBox-v2 (2023) ~\cite{yu2023h2rboxv2}$^2$ & 88.9 & 70.7 & 47.1 & 60.5 & 79.7 & 73.6 & 87.4 & 90.9 & 82.3 & 75.5 & 60.3 & 64.3 & 64.1 & 68.4 & 40.7 & 70.30\\ 
H2RBox-v2 (2023) ~\cite{yu2023h2rboxv2} & 89.1 & 74.6 & 47.5 & 59.8 & 80.7 & 73.1 & 87.8 & 90.9 & 85.6 & 74.1 & 60.7 & 61.7 & 65.8 & 71.8 & 56.0 & 71.96\\ \hline
\rowcolor{gray!20} \multicolumn{17}{l}{$\blacktriangledown$ \textit{Point-supervised OOD}} \\ \hline
\footnotesize Point2RBox (2024) ~\cite{yu2024point2rbox}  & 62.9 & 64.3 & 14.4 & 35.0 & 28.2 & 38.9 & 33.3 & 25.2 & 2.2  & 44.5 & 3.4  & 48.1 & 25.9 & 45.0 & 22.6 & 34.07 \\
\footnotesize Point2RBox+SK (2024) ~\cite{yu2024point2rbox} & 53.3 & 63.9 & 3.7  & 50.9 & 40.0 & 39.2 & 45.7 & 76.7 & 10.5 & 56.1 & 5.4  & 49.5 & 24.2 & 51.2 & 33.8 & 40.27 \\
PointOBB-v3 (2025) ~\cite{zhang2025pointobbv3}  & 30.9 & 39.4 & 13.5 & 22.7 & 61.2 & 7.0 & 43.1 & 62.4 & 59.8 & 47.3 & 2.7 & 45.1 & 16.8 & 55.2 & 11.4 & 41.29 \\
Point2RBox-v2 (2025)~\cite{yu2025point2rbox}$^2$ & 77.9 & 51.6 & 7.5 & 35.3 & 69.6 & 58.5 & 75.1 & 88.3 & 57.6 & 73.1 & 12.3 & 34.1 & 29.6 & 47.2 & 17.8 & 49.04 \\ 
Point2RBox-v2 (2025)~\cite{yu2025point2rbox} & 78.4 & 52.7 & 8.3 & 40.9 & 71.0 & 60.5 & 74.7 & 88.7 & 65.5 & 72.1 & 24.4 & 26.1 & 30.1 & 50.7 & 21.0 & 51.00 \\ 
\hline
\rowcolor{gray!20} \multicolumn{17}{l}{$\blacktriangledown$ \textit{Semi-supervised OOD}} \\ 
\hline
MCL (2025)~\cite{wang2024multi}$^\dagger$ & 88.5 & 79.6 & 46.0 & 65.1 & 80.4 & 81.9 & 87.7 & 90.9 & 78.5 & 85.6 & 57.0 & 68.3 & 66.5 &  74.1 & 54.6 & 73.64\\
Vanilla Baseline$^\dagger$ & 88.5 & 72.9 & 42.4 & 56.3 & 78.4 & 80.3 & 87.2 & 90.8 & 78.3 & 84.1 & 55.4 & 59.8 & 70.7 & 73.7 & 43.3 & 70.80  \\
\hline
\rowcolor{gray!20} \multicolumn{17}{l}{$\blacktriangledown$ \textit{Partial Weakly-supervised OOD (ours)}} \\ 
\hline
\rowcolor{gray!20} \textit{w}/ Partial Horizontal Box$^\dagger$  & 89.0 & 69.3 & 49.7 & 50.6 & 79.3 & 74.4 & 86.2 & 90.9 & 81.9 & 85.2 & 56.2 & 65.1 & 69.1 & 75.8 & 53.4 & 71.74 \\ 
\rowcolor{gray!20} \textit{w}/ Partial Single Point$^\dagger$ & 79.0 & 60.8 & 14.6 & 38.5 & 70.9 & 66.5 & 73.5 & 86.5 & 69.8 & 74.3 & 29.1 & 13.6 & 33.7 & 59.2& 33.8 & 53.59\\ 
\bottomrule

\specialrule{0pt}{2pt}{0pt}
\multicolumn{17}{l}{$^1$PL: Plane, BD: Baseball diamond, BR: Bridge, GTF: Ground track field, SV: Small vehicle, LV: Large vehicle, SH: Ship, TC: Tennis court,} \\
\multicolumn{17}{l}{$\,\;$BC: Basketball court, ST: Storage tank, SBF: Soccer-ball field, RA: Roundabout, HA: Harbor, SP: Swimming pool, HC: Helicopter.} \\
\multicolumn{17}{l}{$^2$Only the training set is used for training. Others without superscripts are trained using the trainval set by default, as in the corresponding published papers.}\\
\multicolumn{17}{l}{$^\dagger$The fully/weakly labeled training data is the training set, and the unlabeled training data is the validation set.} \\
\bottomrule
\end{tabular}}
\end{table*}.

\subsection{Main Results}
\textbf{Partial Horizontal Box:} As illustrated in Table \ref{tab:tab_dota}, our proposed PWOOD framework demonstrates superior performance on the DOTA-v1.5 Dataset-Partial with 10\%, 20\%, and 30\% partial horizontal bounding box annotations, outperforming the SOOD baseline methods that utilize corresponding proportions of rotated bounding box annotations. Specifically, under 10\%, 20\%, and 30\% HBox annotations, PWOOD improves mAP by 3.34\%, 1.08\%, and 0.58\%, respectively, compared to the SOOD baseline method. It means that PWOOD can achieve comparable or even better performance at a lower cost. 
On the DIOR Dataset-Partial in Table \ref{tab_dior}, even when using weakly annotated horizontal boxes, PWOOD achieves performance comparable to the Vanilla Baseline trained with the same proportion of costly rotated box annotations. This further validates that PWOOD delivers equivalent results (54.33\% vs. 54.01\%, 57.89\% vs. 57.07\%, 60.42\% vs. 60.25\% ) at a significantly lower cost compared to high-cost training methods. 

\begin{table*}[htbp]
\begin{minipage}[t]{0.45\columnwidth}
        \caption{mAP comparison on DOTA-v1.5 \textit{val set}.}
    \label{tab:tab_dota}
    
 \resizebox{0.95\textwidth}{!}{
    \begin{tabular}{c l c c c}
        \toprule
        \multirow{2}{*}{\textbf{Task}} & \multirow{2}{*}{\textbf{Method}} & \multicolumn{3}{c}{\textbf{DOTA-v1.5 Dataset-Partial}}\\
        \cmidrule(r){3 - 5} 
         & & 10\% & 20\% & 30\%  \\
        \midrule
        \multirow{2}{*}{WOOD} 
            & H2RBox-v2 & 42.19 & 49.01 & 55.19  \\
            & Point2RBox-v2 & 32.69 & 36.03 & 38.30  \\
        \midrule
        \multirow{7}{*}{\parbox{2.5cm}{\centering SOOD\\(\textit{w}/ Partial RBox)}} 
            & SOOD & 48.63 & 55.58 & 59.23 \\
            & Dense Teacher & 46.90 & 53.93 & 57.86  \\
            & PseCo & 48.04 & 55.28 & 58.03  \\
            & ARSL & 48.17 & 55.34 & 59.02  \\
            & SOOD++& 50.48 & 57.44 & 61.51 \\
            & MCL & 52.61 & 59.63 & 62.63 \\
            & Vanilla Baseline & 49.53 & 58.28 & 61.00  \\
        \midrule
        \multirow{2}{*}{\textbf{PWOOD (ours)}} 
            & \cellcolor{gray!20} \textit{w}/ Partial HBox & \cellcolor{gray!20} 52.87 & \cellcolor{gray!20} 59.36 & \cellcolor{gray!20} 61.58 \\
            & \cellcolor{gray!20} \textit{w}/ Partial Point & \cellcolor{gray!20} 35.33 & \cellcolor{gray!20} 41.54 & \cellcolor{gray!20} 43.02  \\
            		
        \bottomrule
    \end{tabular}}
    \end{minipage}\quad 
\begin{minipage}[t]{0.53\columnwidth}
 \caption{Ablation with different levels of noise adding to HBox/Point annotations on DOTA-v1.0 \textit{val set} and DOTA-v1.5 \textit{val set}. Noise sampled from a uniform distribution $\left[-\sigma H, +\sigma H \right ]$, where $H$ represents the height of the objects, is added to annotations.}
        \label{tab:noise}
\resizebox{\textwidth}{!}{
    \begin{tabular}{c c c c c c c c}
\toprule
\multicolumn{2}{c}{\textbf{Dataset}}  
& \multicolumn{3}{c}{DOTA-v1.0} 
& \multicolumn{3}{c}{DOTA-v1.5} \\
\cmidrule(lr){1-2} \cmidrule(lr){3-5} \cmidrule(lr){6-8}
\multicolumn{2}{c}{\textbf{Noise}}  & 0\% & 10\% & 30\% & 0\% & 10\% & 30\% \\
\midrule

\multirow{2}{*}{\textbf{WOOD}}
& Point2RBox-v2 
& 40.39 & 30.41 & 24.27 
& 36.03 & 24.23 & 21.92 \\

& H2RBox-v2 
& 54.38 & 50.55 & 41.75 
& 49.01 & 44.52 & 35.87 \\

\midrule

\multirow{2}{*}{\textbf{PWOOD}}
& w/ Partial Point 
& 45.01 & 46.34 & 35.50 
& 41.54 & 40.81 & 31.67 \\

& w/ Partial HBox 
& 62.93 & 59.88 & 55.32 
& 59.36 & 56.07 & 51.49 \\

\bottomrule
\end{tabular}
}
\end{minipage}

\end{table*}

Furthermore, we compare our approach with the weakly supervised algorithm H2RBox-v2~\cite{yu2023h2rboxv2}, which only employs partial horizontal bounding box annotations during training. Our model significantly outperforms H2RBox-v2 in the partially weak annotation setting. Specifically, on the DOTA-v1.5 Dataset-Partial, PWOOD achieves large margins of improvement, with gains of 10.68\%, 10.35\%, and 6.39\% for the 10\%, 20\%, and 30\% annotation ratios, respectively. Similarly, on the DIOR Dataset-Partial in Table \ref{tab_dior}, PWOOD demonstrates mAP improvements ranging from 6.56\% to 9.79\%. These results emphasize that PWOOD can effectively mine valid information from unlabeled data, highlighting its efficiency and effectiveness in utilizing limited annotation resources.

\textbf{Partial Single Point:} Given the diversity of weak annotation forms, to validate the robustness of the PWOOD framework under partial weakly supervised settings, we further evaluated its performance in a setting with partial single point. Experimental results in Table \ref{tab:tab_dota} demonstrate that, upon the elimination of scale-related information, the performance of PWOOD experienced a decline; however, when compared to the weakly supervised algorithm with partial single point, Point2RBox-v2~\cite{yu2025point2rbox}, our PWOOD still demonstrated significant superiority. Specifically, on the DOTA-v1.5 Dataset-Partial with 10\%, 20\%, and 30\% partial single point, PWOOD achieved improvements in mAP by 2.64\%, 5.51\%, and 4.72\%, respectively. Under the same annotation ratios on the DIOR Dataset-Partial, it also showed an increase in mAP ranging from 2.36\% to 3.37\%. This demonstrates that our framework exhibits universality across different forms of partially weak annotations.

\textbf{More Results:} We also conducted experiments on DOTA-v1.0/v2.0. As shown in Table~\ref{tab_dior}, PWOOD demonstrates significant performance improvements over WOOD under three different HBox and Point annotation ratios, further proving that PWOOD effectively leverages unlabeled data. Specifically, PWOOD outperforms the Vanilla Baseline across DOTA-v1.0/v1.5/v2.0 datasets under all HBox annotation ratios.
Moreover, As detection difficulty increases, the relative mAP gain of PWOOD over Vanilla Baseline becomes more pronounced. On DOTA-v2.0 which includes more small objects, PWOOD achieves mAP improvements of 6.26\%, 2.36\%, and 2.97\%, respectively, compared to the gain on DOTA-v1.0 (0.89\%, 0.11\%, and 0.45, respectively). This suggests that PWOOD exhibits unique advantages in complex scenes with small objects.

\begin{figure}[tb!]
    
    \begin{minipage}[t]{0.48\columnwidth}
         \includegraphics[width=0.99\linewidth]{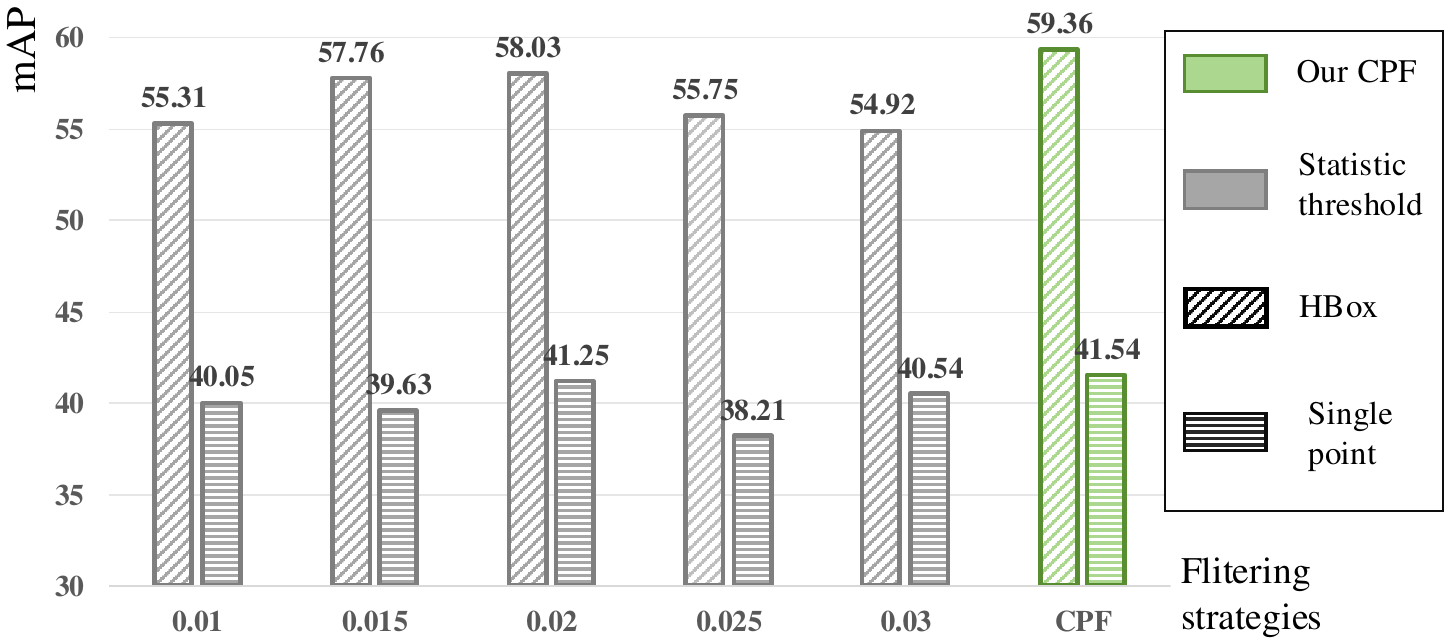}
        \caption{Ablation of pseudo-label filtering strategies.}
        \label{fig:thres_ablation}
    \end{minipage}\quad \quad
    \begin{minipage}[t]{0.48\columnwidth}
          \includegraphics[width=0.99\linewidth]{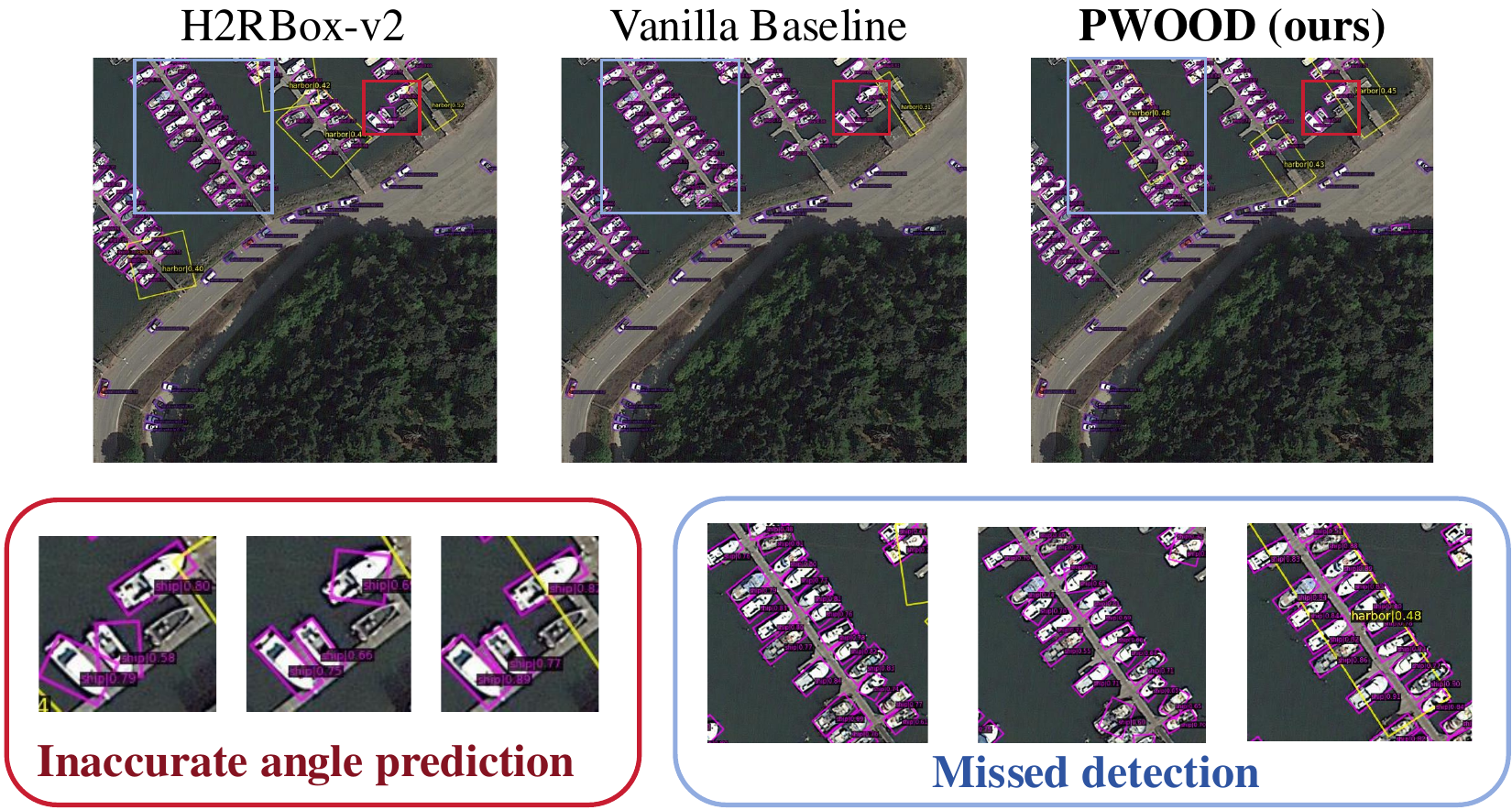}
        \caption{Visualized performance comparison of PWOOD with H2RBox-v2 and the Vanilla Baseline.}
        \label{fig:enter-label}
    \end{minipage}
\end{figure}

What’s more, we evaluate the performance of WOOD, SOOD, and PWOOD on the DOTA-v1.0 \textit{test set} at full scale. Both WOOD, SOOD, and PWOOD are trained on the DOTA-v1.0 \textit{train set}, while WOOD and PWOOD utilize the DOTA-v1.0 \textit{val set} as unlabeled data. As shown in Table \ref{tab:exp_dotav1}, compared to WOOD, PWOOD achieves improvements of 1.44\% (71.74\% vs. 70.30\%) and 4.55\% (53.59\% vs. 49.04\%) under the weak annotation settings of partial HBox and partial single point, respectively, proving that PWOOD fully exploits the potential of unlabeled data. Additionally, compared to the Vanilla Baseline, PWOOD achieves comparable performance, indicating that PWOOD can deliver highly competitive results using low-cost annotations. As illustrated in Figure \ref{fig:enter-label}, the visual comparison among the three frameworks shows that PWOOD exhibits fewer instances of inaccurate angle predictions and missed detections compared to the other two methods. This further validates that PWOOD not only enhances accuracy but also improves robustness in handling complex scenarios. 

\subsection{Ablation Study}

\textbf{Threshold Sensitivity Analysis:} The selection threshold for pseudo-labels is critical to their quality. However, static thresholds often fail to adapt to the specific characteristics of different datasets and training stages, resulting in significant sensitivity of the model to static thresholds. To demonstrate this, we conducted multiple experiments with static thresholds on the DOTA-v1.5 Dataset-Partial with 20\% weak annotations, as illustrated in Figure \ref{fig:thres_ablation}. It is evident that the model's performance varies considerably under different thresholds. Specifically, a slight change in the static threshold from 0.02 to 0.03 leads to a 3.09\% drop in mAP. Furthermore, as shown in Figure \ref{fig:thres_ablation}, the best-performing static threshold achieves a validation mAP of 58.03\% and 41.25\%, respectively, while the introduction of CPF boosts the mAP to 59.36\% and 41.54\%, respectively. This indicates that CPF mechanism effectively improves pseudo-label quality and enhances model performance.

\begin{figure}[t]
    \centering
    \includegraphics[width=0.9\linewidth]{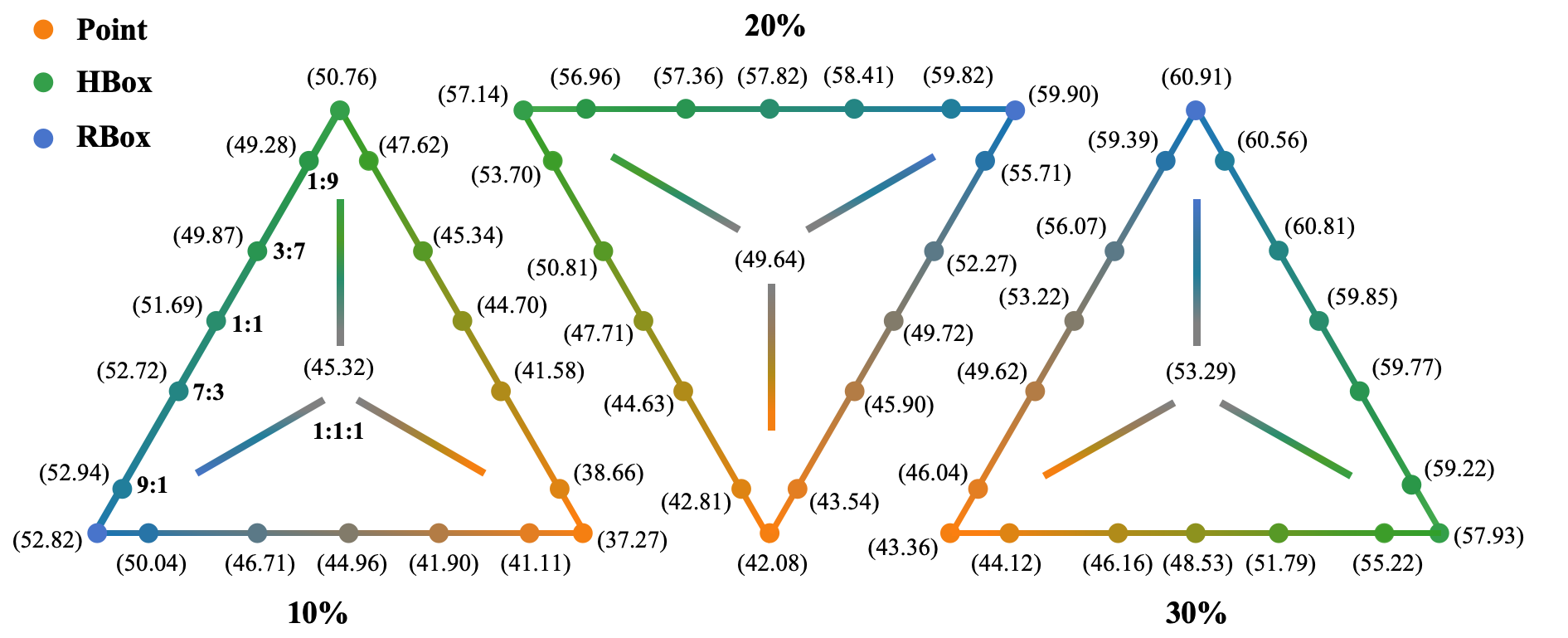}
    \vspace{-8pt}
    \caption{The preliminary experimental results of joint training with various annotations on DOTA-v1.5 Dataset-Partial}
    \vspace{-10pt}
    \label{fig:various}
\end{figure}

\textbf{Robustness Analysis against Noise:} As shown in Table~\ref{tab:noise}, by comparing the performance of WOOD and our PWOOD framework under varying levels of noise interference on the DOTA-v1.5 and DOTA-v1.0 datasets, it is evident that PWOOD exhibits superior robustness compared to WOOD. For example, when 10\% and 30\% noise is added to the DOTA-v1.0 dataset, under 20\% horizontal box supervision, H2RBox-v2 shows mAP drops of 3.83\% and 12.63\%, respectively, while PWOOD only experiences drops of 3.05\% and 7.61\%. This indicates that PWOOD not only achieves superior performance compared to WOOD but also reduces performance degradation under noise interference, as further in Table~\ref{tab:noise}.

\subsection{More Disscussion}
Building upon these demonstrated achievements and considering the inherent diversity of annotation standards in practical applications, we identify a significant opportunity to enhance the PWOOD framework's versatility. Specifically, enabling the framework to support joint training with multi-format labeled data (RBox, HBox, and Point) emerges as a highly promising direction for substantially reducing the challenges associated with training data acquisition. The experimental results are shown in Figure~\ref{fig:various}. Notably, at 20\% multi-format annotations, substituting 10\% of RBox annotations with equivalent proportions of HBoxes results in merely 0.08\% (59.90\% vs. 59.82\%) mAP degradation, respectively. These results demonstrate that our PWOOD successfully bridges the cost-performance gap and provides better cost-accuracy tradeoffs.

\section{Conclusion}
In this paper, we present a new framework, PWOOD, for oriented object detection using a weaker labeling paradigm to further reduce the annotation cost. Within this framework, 
we introduce the orientation and scale learning modules,  endowing the student model with the capability to learn orientation and scale information and resulting in the OS-Student. Moreover, to mitigate the model's sensitivity to static pseudo-label filtering thresholds, 
we propose CPF that dynamically filters pseudo-labels to enhance the model's generalization capability. Our proposed PWOOD model reduces the price in both annotation speed and cost. Extensive experiments on benchmarks DOTA-v1.0/v1.5/v2.0 and DIOR demonstrate that, whether using the partial horizontal box or single point, PWOOD achieves performance comparable to or even surpassing existing WOOD and SOOD algorithms while significantly lowering annotation costs.


\bibliographystyle{plainnat}
\bibliography{references}

\end{document}